\pdfoutput=1
\documentclass[runningheads]{llncs}

\usepackage{accv}

\usepackage{accvabbrv}

\usepackage{amsmath}
\usepackage{amssymb}
\usepackage{graphicx}
\usepackage{booktabs}
\usepackage{multirow}
\usepackage{subcaption}
\usepackage{xcolor}
\usepackage{enumitem}
\usepackage{tcolorbox}
\tcbuselibrary{skins,breakable}

\AtBeginDocument{%
  \setlength{\abovedisplayskip}{3pt plus 1pt minus 1pt}%
  \setlength{\belowdisplayskip}{3pt plus 1pt minus 1pt}%
  \setlength{\abovedisplayshortskip}{1pt}%
  \setlength{\belowdisplayshortskip}{2pt}%
}

\newtcolorbox[list inside=prompt,auto counter,number within=section]{prompt}[1][]{
    breakable, colbacktitle=black!60, coltitle=white, fontupper=\scriptsize,
    boxsep=2pt, left=2pt, right=2pt, top=1pt, bottom=1pt, boxrule=0.5pt,
    before upper={\setlength{\parskip}{1pt}\setlength{\parindent}{0pt}},
    title={#1}, #1,}
\usepackage{hyperref}
\usepackage{orcidlink}

\begin{document}

\title{When, Where, and Why: Hierarchical Causal Explainability
for Video Face Understanding with MLLMs}
\titlerunning{Hierarchical Causal Explainability for Video Face Understanding}

\author{Anubhav Kataria\inst{1} \and
Shreya Ghosh\inst{2} \and
Jingwen Ye\inst{1} \and
Abhinav Dhall\inst{1}}
\authorrunning{A.~Kataria et al.}
\institute{Monash University, Melbourne, Australia \and
Indian Institute of Technology Kharagpur, Kharagpur, India}

\maketitle

\begin{abstract}
Understanding facial behaviour in video requires answering three fundamental questions: \emph{when} a key facial expression occurs, \emph{where} the model attends on the face at that moment, and \emph{why} that evidence is necessary rather than merely correlated. Today's multimodal large
language models (MLLMs) can label and describe video, but they rarely provide faithful explanations,
especially for small facial movements that change over time. We present
\textit{\textbf{VideoFace-X-LLM}}, a framework for explainable facial video understanding comprising two key components. First, we introduce \emph{ExplainFace}, a weakly supervised video-text corpus for facial behaviour understanding, automatically constructed from CelebV-HQ videos and validated through human evaluation and a two-model audit (at most 2.3\% contradictions), demonstrating consistent gains across two backbone architectures. Second, H-TCC (Hierarchical Temporal Spatial Causal Concepts), which
finds the key temporal moments (\emph{When}), the facial regions (\emph{Where}), and verifies its causal necessity (\emph{Why}) through targeted occlusion. Fine-tuning on ExplainFace
raises appearance Macro F1 from $19.48\%$ to $32.87\%$ ($+13.39$), with gains of $+4.38$ on
actions, $+6.77$ on temporal emotion, and $+11.84$ F1 on deepfake detection. We also \emph{measure} the explanations pointing lift, face/edge lift, stability, an Adebayo
sanity check, and a necessity test rather than only showing examples. Our models, code, and corpus are available on request for research use.

\end{abstract}

\begin{figure}[t]
  \centering
  \includegraphics[width=0.95\textwidth]{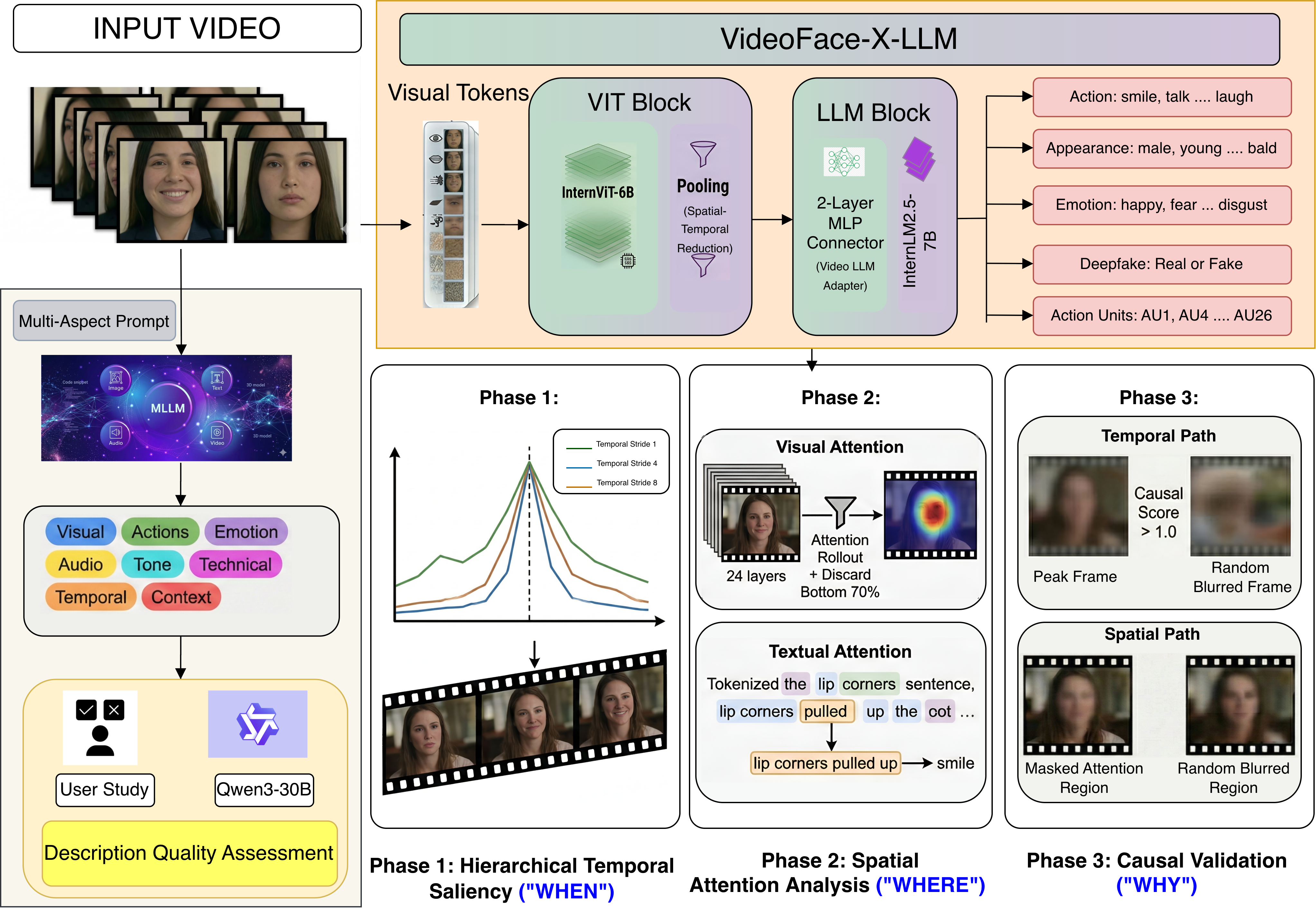}
  \caption{Overview of VideoFace-X-LLM. \textbf{Left:} the ExplainFace corpus is built by
  prompting Gemini for multi-aspect video descriptions, then auditing them for quality and
  bias. \textbf{Top:} the MLLM is fine-tuned on this corpus. \textbf{Right:} H-TCC explains
  each prediction by finding \emph{when} the key moment occurs, \emph{where} the model looks
  on the face, and \emph{why} that evidence is needed.}
  \label{fig:teaser}
  \vspace{-8mm}
\end{figure}

\section{Introduction}
\label{sec:intro}

\textit{Faces in motion} are one of the richest signals of human behaviour. A raised brow
becomes surprise only when we see it change over time; a smile becomes warmth or sarcasm
depending on the gaze, head motion, voice, and context around it. Such behaviour is
temporal, multimodal, and causal: meaning comes not from single frames, but from how visual
and audio signals change together. To understand it well, a model should answer three
questions: \emph{when} a key event happens, \emph{where} it shows on the face at that moment,
and \emph{why} that region is needed and not just present by chance. The three are linked:
a good explanation must find the right moment and region, then confirm that this evidence
drives the output.

Multimodal large language models (MLLMs) have improved fast on general vision language
tasks, but they are still weak at fine-grained facial behaviour. Existing
MLLMs~\cite{geminiteam2025gemini25pushingfrontier,yang2025qwen3} are trained on broad data
and lack the priors needed to read small facial motions, micro-expression changes, and cues
that depend on time but they remain strong at open-ended video description and soft-labeling. This is even harder for explanation, where a model must justify a
prediction \emph{over time}, not just describe a still image.

Recent models such as InternVideo2.5~\cite{wang2025internvideo2},
GPT-4o~\cite{openai2024gpt4osystemcard}, and Qwen3-VL~\cite{yang2025qwen3} are strong at
captioning, VQA, and general video analysis, yet they struggle with human-centric reasoning.
Their training data is broad but thin on detailed labels for expressions, affect, body
language, and temporal change. The real problem is not the models but the data: there is no large video-text dataset that describes faces in detail and lines up those descriptions with time. Weakly supervised, language-model-based dataset building has been
tried for still images (for example FaceLLM's FairFaceGPT~\cite{shahreza2025facellm}), but
not for video. Moving from images to video is harder, because it needs grounding of facial
attributes in both space and time, plus reasoning over small changes that unfold over time.

We address this with \textit{VideoFace-X-LLM}, a model for explainable facial behaviour in
video. We build \textit{ExplainFace}, a weakly supervised video text corpus generated with
Gemini~\cite{geminiteam2025gemini25pushingfrontier}, giving time-aligned descriptions of
facial attributes, expression dynamics, and audio visual context. To explain predictions,
we introduce H-TCC (Hierarchical Temporal Spatial Causal Concepts), which finds
\textit{when} key events occur, \textit{where} the model looks on the face, and \textit{why}
that evidence is needed, using hierarchical temporal saliency, spatial attribution, and a
targeted occlusion test that confirms the identified evidence is necessary for the prediction.
Importantly, fine-tuning on ExplainFace helps \emph{two} different backbones, and we
\emph{measure} the explanations rather than only showing examples. Our main contributions are:
\begin{itemize}[leftmargin=1.2em, itemsep=2pt]
\item \textbf{ExplainFace: a checked, weakly-labeled dataset.} We label CelebV-HQ videos
using Gemini instead of expensive frame-by-frame human labeling. We test how good these
labels are with two human studies and a two-model check for bias, and show the labels
still work well when used with two different model backbones.
\item \textbf{H-TCC: a when/where/why framework that we actually measure.} Every step
of H-TCC is defined with clear equations, and we score the explanations with quantitative
faithfulness metrics (pointing/face/edge lifts, stability, a sanity check, and a necessity
test), not just example cases.
\item \textbf{VideoFace-X-LLM.} A video-understanding model for faces that explains its
answers by pointing to the exact moments and regions that led to its decision.
\end{itemize}

\section{Related Work}
\label{sec:related}
\noindent \textbf{MLLMs for Vision Language Understanding.} Recent MLLMs such as
GPT-4o~\cite{openai2024gpt4osystemcard}, InternVideo2.5~\cite{wang2025internvideo2},
LLaVA-NeXT~\cite{zhang2024llava}, and Qwen3-VL~\cite{yang2025qwen3} have pushed general
image and video reasoning, but lack the domain priors and time-aligned supervision needed
for subtle facial behaviour, motivating our specialised corpus and adapted MLLM.

\noindent \textbf{Human-Centric Facial Behaviour Understanding.}
Many works apply MLLMs and foundation models to faces: recognition~\cite{deng2019arcface},
attribute analysis~\cite{liu2015deep}, understanding~\cite{chaubey2026face}, and multimodal
reasoning~\cite{cheng2024emotion}. A recent survey~\cite{shahreza2025foundation} reviews how
foundation models and LLMs are used for biometric security. Beyond CLIP, self-supervised methods learn facial
representations, such as the masked-autoencoder MARLIN~\cite{cai2023marlin},
FaRL~\cite{zheng2022general}, DINO-style distillation~\cite{bulat2022pre}, and
ComFace~\cite{akamatsu2025comface}. Multi-task models~\cite{narayan2025facexformer} jointly
do face parsing, landmarks, pose, and attributes. HumanOmni~\cite{zhao2025humanomni} uses a
dedicated face branch and does well on expression and emotion. Still, no existing work gives both time-aligned video labels and clear, trustworthy explanations which is what \textit{VideoFace-X-LLM} sets out to do.

\noindent \textbf{Bias and Explainability for Facial Behaviour Understanding.} Facial
datasets carry demographic imbalance that flows into predictions, with uneven results across
gender, race, and age~\cite{dominguez2024metrics}. LLMs can amplify such bias when describing
faces~\cite{gallegos2024bias}, and bias checks for video face descriptions are still missing.

\noindent \underline{\textit{1) Temporal Explainability (\emph{When}).}}
Several methods~\cite{zeng2024timesuite,yu2023sevila,huang2024vtimellm} find key moments in
video. TimeSuite~\cite{zeng2024timesuite} adds temporal grounding,
SeViLA~\cite{yu2023sevila} picks keyframes, and VTimeLLM~\cite{huang2024vtimellm} and
TimeChat~\cite{ren2024timechat} encode timestamps as text.
Grounded-VideoLLM~\cite{wang2025groundedvideollm} splits video into spatial and temporal
tokens. However, looking at only one time scale can miss events that overlap.

\noindent \underline{\textit{2) Spatial Explainability (\emph{Where}).}}
Grad-CAM~\cite{selvaraju2017gradcam} makes spatial heatmaps from gradients but does not suit
transformer encoders or video. For vision transformers, attention
rollout~\cite{abnar2020quantifying} traces information across layers, with refinements that
mix gradients and attention~\cite{chefer2021transformer}. GMAR~\cite{jo2025gmar} reweights
heads by gradient importance, and LVLM-Interpret~\cite{stan2024lvlminterpret} extends rollout
to large models. However, these maps do not line up with time.

\noindent \underline{\textit{3) Causal Validation (\emph{Why}).}}
Occlusion methods test whether evidence is needed by masking inputs and watching the output
change. Zeiler and Fergus~\cite{zeiler2014visualizing} used systematic occlusion for CNNs,
RISE~\cite{petsiuk2018rise} uses random masks, and Fong and Vedaldi~\cite{fong2017interpretable}
frame it as finding the smallest change that flips the output. The problem is that they
perturb blindly, without knowing \emph{when} or \emph{where} to look, which is slow and hard
to read. They are also not built for video MLLMs, where masks must cover both time and space.

Existing methods usually answer only one of \emph{when}, \emph{where}, or \emph{why}, each
with limits that our H-TCC addresses together.

\section{ExplainFace Corpus}
\label{sec:ExplainFaceCorpus}
Most MLLMs~\cite{wang2025internvideo2,yang2025qwen3} are trained on broad data with little
human-centric detail, so they miss subtle facial behaviour and affect over time. We close
this gap with \textit{ExplainFace}, a video text corpus for detailed analysis of faces over time. We use videos from CelebV-HQ~\cite{zhu2022celebv} and generate descriptions with
the Gemini API~\cite{geminiteam2025gemini25pushingfrontier}. The pipeline has two
stages: a system prompt for video analysis (Prompt 3.1) followed by a multi-aspect
description prompt; then a bias audit of those descriptions (Prompt 3.2). \cref{fig:teaser}
shows the pipeline on the left.

\subsection{Prompting for ExplainFace Corpus}
\noindent \textbf{System Prompt.}
Because the model can answer in different ways for different
videos, we use a fixed system prompt (Prompt 3.1) so the output stays detailed and
structured.

\noindent \textbf{Multi-Aspect Description Generation.} For each video we generate
descriptions over eight aspects: \textit{theme \& narrative, visual elements, actions \& movements, audio components, emotional tone, technical aspects, temporal flow, and context \& relationships}. These cover both still details and change
over time. We use markdown headings so the format is consistent. Unlike
FaceLLM~\cite{shahreza2025facellm}, which uses FairFace metadata to guide generation, we give
Gemini no prior information about the video, because temporal content is hard to summarise in
metadata. We let the model watch the video and extract the details itself.
\begin{prompt}[title={Prompt \thetcbcounter: System Prompt for Video Analysis},float=t, unbreakable]
\small
\textbf{System:} You are an AI assistant specialized in analyzing video content with high
accuracy and detail, suitable for multimodal video understanding and emotional analysis
applications. Your task is to analyze and describe videos comprehensively, including visual
elements, audio components, temporal dynamics, and emotional content. As an assistant, you
should always analyze given video content thoroughly and provide objective, detailed
descriptions based on observable information.

Do not make assumptions beyond visible and audible traits. When analyzing a video, consider
visual elements (scenes, objects, people, actions, lighting, camera movements, visual
effects), audio components (speech, music, sound effects, ambient noise), temporal dynamics
(sequence of events, transitions, pacing, duration), emotional expressions and behaviors
(facial expressions, body language, vocal tone, emotional states), contextual information
(setting, relationships between subjects, narrative flow), and technical aspects (video
quality, editing techniques, production values).

I may provide you with additional context about the video content, duration, or specific
aspects to focus on. If you are asked for specific features or attributes, only describe what
is asked in the question. Never say ``I'm unable to analyze''   instead, provide detailed
descriptions based on the visual and auditory information available. Your answers should be
comprehensive text descriptions (one or multiple paragraphs) that paint a clear picture of
the video's content and emotional landscape.
\end{prompt}
\noindent \textbf{Linking aspects to tasks.} Each description aspect supervises a downstream
task, as shown in \cref{tab:aspect_task}. For example, ``Visual Elements'' teaches
appearance, ``Actions'' teaches action labels, and ``Emotional Tone'' teaches emotion and
Action Units. This is why a general description corpus can train fine-grained facial tasks.

\begin{table}[b]
\centering
\caption{How each ExplainFace description aspect supervises a downstream task.}
\label{tab:aspect_task}
\scalebox{0.9}{
\begin{tabular}{l l}
\toprule
\textbf{Description aspect} & \textbf{Task it mainly supervises} \\
\hline \hline
Visual Elements                & Appearance attributes \\
Actions \& Interaction         & Action recognition \\
Emotional Tone \& Expressions  & Emotion recognition, Action Units \\
Temporal Flow                  & Temporal emotion, deepfake (motion cues) \\
Audio Components               & Multimodal context (speech, affect) \\
Technical Aspects              & Manipulation / deepfake artifacts \\
\bottomrule
\end{tabular}}
\vspace{-8mm}
\end{table}

\noindent \textbf{Quality Analysis for ExplainFace Corpus.} AI-generated text can
hallucinate, so we check the corpus with a `human-in-the-loop' study. To test whether humans
can match the AI descriptions to the videos, we ran two user studies with $10$ participants
who have computer-vision experience. The authors did not take part\footnote{All procedures
followed the affiliated University's Human Research Ethics Committee approval.}.

\noindent \underline{\textit{User study 1 (Description quality):}}
We randomly pick videos from the ExplainFace corpus and split them across
10 participants. After dropping incomplete answers, 531 pairs are left. Each participant gets the same instructions as Gemini and writes
their own eight-aspect description, without seeing Gemini's text. We then compare human and
Gemini descriptions with semantic and lexical metrics. The goal is to check whether humans
and Gemini describe the \emph{same content}, not the same words.

\noindent \underline{\textit{User study 1 Results:}} \cref{tab:user_study1} compares human and
Gemini descriptions across $531$ videos and eight aspects, using SBERT~\cite{reimers2019sentence} cosine similarity for
meaning and BLEU~\cite{papineni2002bleu}/ROUGE~\cite{lin2004rouge} for word overlap. Semantic similarity is high while lexical overlap is
low: SBERT averages $0.45$, against BLEU-1 $0.06$ and ROUGE-L $0.14$. For reference, two humans describing the \emph{same} video reach only $0.38$, and a human and
Gemini describing \emph{different} videos $0.30$, so $0.45$ exceeds both. This is exactly what we
want: it shows humans and Gemini say the \emph{same thing in different words}. Gemini writes
longer, structured text; humans write short, informal notes. The pattern holds across
aspects: Technical Aspects is highest (SBERT $0.54$, shared vocabulary), while Emotional Tone
($0.44$) and Context ($0.33$) are lower, since they are more subjective.

\begin{table}[t]
\centering
\caption{User Study 1: human vs.\ Gemini descriptions ($531$ videos). SBERT measures meaning;
BLEU and ROUGE measure word overlap. High SBERT with low BLEU/ROUGE means both describe the
same content in different words.}
\label{tab:user_study1}
\scalebox{0.85}{
\begin{tabular}{l c c c}
\toprule
\textbf{Aspect} & \textbf{SBERT} $\uparrow$ & \textbf{BLEU-1} $\uparrow$ & \textbf{ROUGE-L} $\uparrow$ \\
\hline \hline
Theme \& Narrative       & 0.46 & 0.06 & 0.13  \\
Visual Elements          & 0.43 & 0.05 & 0.14  \\
Actions \& Movements     & 0.46 & 0.06 & 0.14  \\
Audio Components         & 0.45 & 0.08 & 0.17  \\
Emotional Tone           & 0.44 & 0.04 & 0.11  \\
Technical Aspects        & 0.54 & 0.06 & 0.16  \\
Temporal Flow            & 0.46 & 0.07 & 0.14  \\
Context \& Relationships & 0.33 & 0.04 & 0.10  \\
\midrule
\textbf{Overall}         & \textbf{0.45} & \textbf{0.06} & \textbf{0.14}  \\
\bottomrule
\end{tabular}}
\vspace{-8mm}
\end{table}

\noindent \underline{\textit{User study 2 (Video retrieval from description):}} Here we test
whether the descriptions are detailed enough to point to one exact video. We pick 1000 descriptions and split them across 10 participants. After dropping incomplete trials, 917 are left. Each participant reads a description and
picks the matching video from four candidates. If a description is accurate and specific, the
participant should beat the $25\%$ chance baseline. We use three difficulty levels. \textit{i)
Easy:} the three wrong candidates are random, so they usually show different scenes; this
tests if the description is factually correct. \textit{ii) Medium:} the wrong candidates are
chosen to be highly similar in description embedding space (for example, all show a person
speaking to camera), so a generic line like `a man talks' is not enough; the description must
capture clothing, expression, or background. \textit{iii) Hard:} all four candidates are
different time segments of the \emph{same} video (using CelebV-HQ metadata). They share the
same context, so the only way to pick the right one is fine temporal detail: which
expression appears when, and in what order events happen.

\noindent \underline{\textit{User study 2 Results:}} \cref{tab:user_study2} shows $99.18\%$ on
easy, $94.55\%$ on medium, and $91.24\%$ on hard, all far above the $25\%$ baseline. Even on
the hard, same-clip task, people pick correctly $91\%$ of the time, which shows the
descriptions capture temporal detail: expression changes, emotional shifts, and event order.

\begin{table}[t]
\centering
\caption{User Study 2: human video retrieval accuracy. A participant reads a description and
picks the matching video from four options. Chance accuracy is $25\%$.}
\label{tab:user_study2}
\begin{tabular}{l| l| c c}
\toprule
\textbf{Level} & \textbf{What varies} & \textbf{Samples}  & \textbf{Human Acc.} \\
\hline \hline
Easy & Random videos & 368 & 99.18\% \\
Medium & Similar content & 275  & 94.55\% \\
Hard & Same-clip segments & 274  & 91.24\% \\
\hline
Overall &   & 917 & 95.42\% \\
\bottomrule
\end{tabular}
\vspace{-8mm}
\end{table}

\subsection{Bias Analysis of ExplainFace Corpus}
MLLM descriptions can be biased, often through stereotypes or guesses from
appearance~\cite{dominguez2024metrics}. So we add a bias-audit stage that checks each
description. A separate prompt asks the model to flag biased language and suggest a neutral
version grounded in what is visible. This makes the descriptions both detailed and objective.

\noindent \textbf{Subjectivity Audit.} Descriptions can go beyond what is visible. A model
might call a neutral face \textit{`reserved'}, or read \textit{`frustration'} from a raised
brow. These are not wrong, but they are interpretations shaped by priors. To measure how much
interpretive language is in ExplainFace, we use a two-stage audit.

\noindent \underline{\textit{Stage 1:}} A prompt (Prompt 3.2) reviews each description with its
video. For every interpretive statement, the model returns: (i) the exact phrase, (ii) why it
goes beyond what is seen, and (iii) a more grounded alternative. If nothing is flagged, it
says the description is purely observational.

\noindent \underline{\textit{Example flagged pair.}} For one clip, the draft said the person
``looks \emph{aggressive}.'' The audit flagged this as a guess about character and suggested
``the brows are drawn together and the jaw is tense''  - keeping the visible cue but dropping
the assumption.

\noindent \underline{\textit{Observations from Stage 1:}} We run the audit with two models:
\textit{Gemini 2.5 Flash}~\cite{geminiteam2025gemini25pushingfrontier} and
\textit{Qwen3-30B-Instruct}~\cite{yang2025qwen3}. Gemini calls about $70\%$ of descriptions
observational and $30\%$ interpretive; Qwen3 calls about $60\%$ observational and $40\%$
interpretive. The gap shows the task is fuzzy: even two LLMs disagree on where observation
ends and interpretation begins. \textit{This supports our view that objectivity in facial
description is a spectrum, not a yes/no.}

\noindent \underline{\textit{Stage 2:}} Next we compare each flagged phrase with its suggested
alternative. The two should keep the same meaning but change the framing. To test this, we use
four metrics: \textit{i) Cosine similarity} on \textit{MiniLM-L6-v2}
embeddings~\cite{wang2021minilmv2}; \textit{ii) BERTScore}~\cite{zhang2020bertscore} for meaning (should stay high);
\textit{iii) BLEU}~\cite{papineni2002bleu} for word overlap (should drop if wording changed); and \textit{iv) NLI}
with DeBERTa~\cite{he2020deberta} to label each pair as entailment, neutral, or contradiction
(high entailment means the revision follows from the original).

\begin{prompt}[title={Prompt \thetcbcounter: Bias Analysis},float=t, unbreakable]
\small
\textbf{System:} You are an AI ethics assistant. Review the following video and its
accompanying text description for potential biases.

\textbf{Human:} Does the description use stereotypical language related to gender, ethnicity,
or age? Does it make unwarranted assumptions about emotion, intent, or character based on
appearance?

Your task is to:
\begin{itemize}[leftmargin=1em, itemsep=0pt, topsep=0pt, parsep=0pt]

    \item Analyze both the video and the text description provided below.
    \item Identify any specific phrases or sentences in the description that could be
    considered biased, stereotypical, or make unsupported assumptions.
    \item For each identified statement, explain \textit{why} it is potentially problematic.
    \item Suggest a more neutral, objective alternative for each flagged statement, grounding
    your suggestion in observable facts from the video.
\end{itemize}
If no biased statements are found, state that the description appears to be objective and free
of bias. Your response \textbf{must} be structured using the exact following markdown headings:

\textbf{Overall Assessment} \textit{(Provide a one-sentence summary of your findings.)}

\textbf{Bias Analysis} \textit{(If biased statements are identified, list each one as follows:)}\\
\textbf{Problem Statement 1:} [The exact quote from the description]\\
\textbf{Explanation 1:} [Why it is potentially problematic]\\
\textbf{Suggested Alternative 1:} [A more neutral, objective rephrasing]

\textbf{Final Conclusion} \textit{(A brief concluding thought on the description's overall objectivity.)}

\textbf{Video Description to Analyze:}\\
\{description\_text\}\\
\textbf{AI:} \{Structured bias analysis with overall assessment, identified issues (if any),
and final conclusion.\}
\end{prompt}

\begin{table}[b]
\centering
\caption{Semantic similarity between flagged (interpretive) statements and their suggested
observational alternatives. Semantic: meaning; Lexical: word overlap; NLI: Natural Language
Inference; Ent: Entailment, Neu.: Neutral, Con.: Contradiction.}
\label{tab:subjectivity_metrics}
\scalebox{0.75}{
\begin{tabular}{l || c c || c c || c c c}
\toprule
& \multicolumn{2}{c||}{\textit{Semantic $\uparrow$}} & \multicolumn{2}{c||}{\textit{Lexical $\downarrow$}} & \multicolumn{3}{c}{\textit{NLI (\%)}} \\
\textbf{Auditor}  & \textbf{BERT} & \textbf{Cosine} & \textbf{BLEU} & \textbf{ROUGE-L} & \textbf{Ent.} $\uparrow$ & \textbf{Neu.} & \textbf{Con.} $\downarrow$ \\
\textbf{ (Pair)} & \textbf{Score} & \textbf{Sim.} & &  & &  & \\
\hline \hline
\begin{tabular}{c} Gemini \\ 31,128 \end{tabular}& 0.93 & 0.73 & 0.21 & 0.48 & 53.8 & 43.9 & \textbf{2.30} \\
\begin{tabular}{c} Qwen3 \\ 33,008 \end{tabular} & 0.93 & 0.74 & 0.20 & 0.46 & 36.9 & 61.5 & \textbf{1.60} \\
\bottomrule
\end{tabular}}
\vspace{-8mm}
\end{table}

\noindent \underline{\textit{Observations from Stage 2:}} \cref{tab:subjectivity_metrics} reports
two auditors over $31{,}128$ and $33{,}008$ pairs. When a flagged sentence is rewritten to be
more objective, the meaning stays the same: BERTScore F1 is $0.93$ for both Gemini and Qwen3.
At the same time, BLEU is low ($0.21$ and $0.20$), so the wording changed a lot but the
meaning did not. The audit changes \emph{how} something is said, not \emph{what} is said.
Contradiction rates are very low: only $2.3\%$ (Gemini) and $1.6\%$ (Qwen3), so over $97\%$ of
pairs stay factually consistent. Reaching this at scale, with no human annotation, shows the
corpus is robust. Neutral cases ($43.9\%$ for Gemini, $61.5\%$ for Qwen3) are usually extra
hedging or detail, not errors. The two auditors disagree on how much is subjective, yet they
give almost the same similarity and contradiction numbers, so our conclusion does not depend on
the auditor. Together, this shows ExplainFace is accurate and rich, and its interpretive
language links low-level cues to human-level meaning.

\noindent \underline{\textit{Final Conclusion:}} After the audit we keep the full corpus,
including interpretive descriptions: facial behaviour is naturally subjective, and this
interpretive layer matters for emotion and attribute prediction removing it would leave
only anatomy. As our results show, these descriptions help rather than hurt.

\section{VideoFace-X-LLM}
We propose \textit{VideoFace-X-LLM}, a video language model fine-tuned for facial video
understanding. We build on InternVideo2.5~\cite{wang2025internvideo2} and adapt it to the face
domain by fine-tuning on ExplainFace with \textit{Fully Sharded Data Parallel} (FSDP)
training. \cref{fig:teaser} shows the architecture.

\noindent \textbf{Base Model Details.} InternVideo2.5~\cite{wang2025internvideo2} has three
parts: a vision encoder, a vision language connector, and a language decoder. The vision encoder follows InternViT: $24$ transformer layers, hidden size
$1024$, $16$ heads. It reads frames at $448\times448$ with patch size $14$, and uses Flash
Attention~\cite{dao2023flashattention} for speed. The connector is a two-layer MLP with $0.5$
downsampling that maps visual features into the language space. The decoder is
InternLM2.5-7B~\cite{cai2024internlm2}: $32$ layers, hidden size $4096$, group-query attention
with $32$ heads and $8$ key/value heads.

\noindent \textbf{Fine-tuning Details.} We train on the question answer pairs from ExplainFace
(\cref{sec:ExplainFaceCorpus}) with the matching CelebV-HQ videos. We fine-tune end-to-end to
predict the answer, using FSDP~\cite{zhao2023pytorch}.

\section{Explainability: \underline{H}ierarchical \underline{T}emporal-Spatial \underline{C}ausal \underline{C}oncepts (H-TCC)}
\label{sec:htcc}
\begin{figure}[t]
    \centering
    \includegraphics[width=0.95\linewidth]{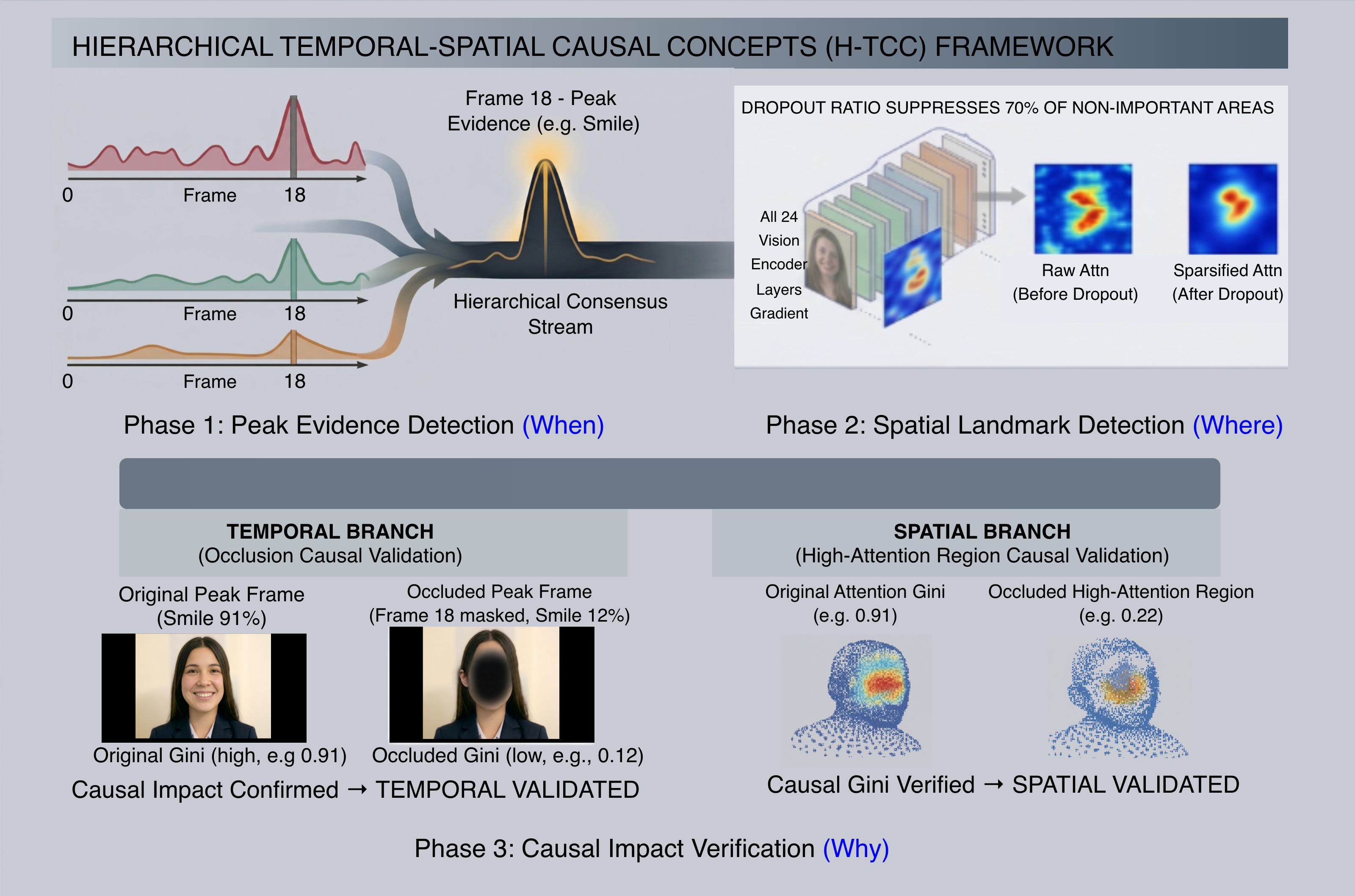}
    \caption{Overview of H-TCC. Phase 1 detects the peak moment (\emph{When}); Phase 2 localises
    where the model attends on the face (\emph{Where}); Phase 3 tests whether that evidence is needed (\emph{Why}).
    See \cref{sec:htcc}.}
    \label{fig:htcc}
    \vspace{-8mm}
\end{figure}

Video language models~\cite{geminiteam2025gemini25pushingfrontier,yang2025qwen3} are strong,
but their deep encoders and cross-modal attention make it hard to say which region or moment
drove a prediction. We address this with H-TCC, which finds \emph{when} a key event happens,
\emph{where} the model looks, and \emph{why} that evidence is needed (\cref{fig:htcc}).

\noindent\textbf{Problem definition.} Given a video $V=\{f_1,\dots,f_T\}$ and a predicted
attribute $a$, H-TCC returns a temporal peak $t^\star$ (\emph{When}), a spatial attention map
$R$ over image patches (\emph{Where}), and a necessity decision (\emph{Why}). We call an
explanation \emph{faithful} if perturbing the identified evidence changes the prediction more
than perturbing random evidence.

\noindent \textbf{Phase 1: Hierarchical Temporal Saliency (When).} Phase 1 finds the moments
that matter most for a prediction. First, we measure how strongly the 
model supports a target
label $a$, using a margin between $a$ and its best competing label:
\begin{equation}
m \;=\; g(a) \;-\; \max_{c \neq a} g(c) ,
\label{eq:margin}
\end{equation}
where $g$ is the model's score for a label. For appearance and action, $g$ is a binary Yes/No
presence probe, so $m = \max_{i\in\mathrm{Yes}} z_i - \max_{j\in\mathrm{No}} z_j$ over the
next-token logits $z$. For emotion, $g(c)$ is the length-normalised sequence log-likelihood of
label $c$ under the carrier ``the dominant emotion is $c$'',
$g(c)=\frac{1}{|c|}\sum_k \log p\!\big(y^{c}_k \mid y^{c}_{<k}, V\big)$, and the competitors are
the other emotion classes. We use a margin, not a probability, because it does not saturate
near $1.0$ and so still moves when evidence is removed. Next, for each frame $f$ with
visual tokens $\mathcal{T}_f$ (embeddings $\mathbf{e}_t$), we measure how much the frame pushes
the attribute logit $y_a$, using the gradient times the token value:
\begin{equation}
s_f \;=\; \sum_{t\in\mathcal{T}_f}\Big|\big\langle \nabla_{\mathbf{e}_t} y_a,\;\mathbf{e}_t\big\rangle\Big| .
\label{eq:frameimp}
\end{equation}
We then smooth these scores at three time scales (strides $1,4,8$), normalise each, and
multiply them. A frame becomes a peak only if it is important at \emph{all} three scales, which
removes noise:
\begin{equation}
\hat{s}^{(r)}=\mathrm{norm}\big(\mathrm{AvgPool}_r(s)\big),\qquad
C_f=\!\!\prod_{r\in\{1,4,8\}}\!\!\hat{s}^{(r)}_f,\qquad
t^\star=\arg\max_f C_f .
\label{eq:consensus}
\end{equation}
We also report how stable the saliency is across scales, as the mean pairwise Spearman
correlation $\rho_S$ between the per-stride curves:
\begin{equation}
\mathrm{Stab}=\frac{1}{|\mathcal{R}|}\sum_{r<r'}\rho_S\big(\hat{s}^{(r)},\hat{s}^{(r')}\big).
\label{eq:stab}
\end{equation}
To confirm the saliency comes from learned knowledge, not a fixed pattern, we apply cascading
model randomization~\cite{adebayo2018sanity}: we randomize the top-$k$ decoder layers
cumulatively, recompute the consensus $C^{(k)}$, and correlate it with the original. A faithful
map collapses to about zero as $k$ grows:
\begin{equation}
\rho_k=\rho_S\big(C^{(0)},C^{(k)}\big),\qquad \rho_0=1,\qquad \rho_k\to 0 .
\label{eq:adebayo}
\end{equation}

\noindent \textbf{Phase 2: Spatial Attention Analysis (Where).} Phase 2 shows where the model
looks at the peak moment. We give two views: \textit{visual attention} from the vision encoder
and \textit{textual attention} from the language model. \underline{\textit{Visual Attention:}}
we read attention from all $24$ encoder layers at the peak frame (each emotion segment is
handled separately). For each layer we average the heads, zero the lowest $70\%$ of values
(those at or below the $0.7$ quantile $\theta^{(l)}_{0.7}$, to remove edge noise), add a
residual, normalise, and multiply across layers~\cite{abnar2020quantifying}:
\begin{equation}
\hat{A}^{(l)}_{ij}=A^{(l)}_{ij}\,\mathbb{1}\!\big[A^{(l)}_{ij}>\theta^{(l)}_{0.7}\big],\quad
\tilde{A}^{(l)}=\mathrm{rownorm}\big(\hat{A}^{(l)}+I\big),\quad
R=\Big(\textstyle\prod_{l=L}^{1}\tilde{A}^{(l)}\Big)_{0,\,1:}.
\label{eq:rollout}
\end{equation}
The $32\times32$ map $R$ is the class-token row over the patch tokens, normalised by its max,
and overlaid on the frame. We localise the face with MTCNN (Multi-Task Cascaded
CNN)~\cite{zhang2016mtcnn}, or a central crop if no face is found. With $N{=}1024$ patches, face region $F$, and a two-patch-wide strip $E$ along one image border ($|E|{=}2\times32{=}64$), we measure whether attention concentrates on the face and avoids the background border, against a uniform baseline:
\begin{equation}
\mathrm{FaceLift}=\frac{\sum_{p\in F}R_p\big/\sum_p R_p}{|F|/N},\qquad
\mathrm{EdgeLift}=\frac{\sum_{p\in E}R_p\big/\sum_p R_p}{|E|/N}.
\label{eq:lift}
\end{equation}
We also run a pointing game: does the most-attended patch fall on the face, above chance?
\begin{equation}
\mathrm{PointLift}=\frac{\tfrac{1}{K}\sum_{k}\mathbb{1}\!\big[\arg\max_p R^{(k)}_p\in F\big]}{|F|/N}.
\label{eq:point}
\end{equation}
Here $\mathrm{FaceLift},\mathrm{PointLift}>1$ mean the model looks at the face, and
$\mathrm{EdgeLift}<1$ means it avoids borders. \underline{\textit{Textual Attention:}} we read
the model's natural-language explanation, run symmetric rollout over the language layers, and
find attribute-specific phrases from a small evidence vocabulary. Their influence on the
attribute token gives a ranked list of supporting evidence. For emotion, we do this per
segment with segment-specific text.

\noindent \textbf{Phase 3: Necessity Validation via Targeted Occlusion (Why).} The first two
phases tell us \emph{when} and \emph{where} the model's evidence lies; Phase 3 confirms that
this evidence is actually \emph{necessary} for the prediction, rather than merely present
alongside it. We test necessity directly through targeted occlusion: we apply a strong Gaussian
blur to the specific evidence identified by Phases 1 and 2, and use VideoFace-X-LLM itself to
check whether the target attribute is still detected after masking. Because the evidence is already localised in time and space,
the test is far cheaper and easier to read than blind occlusion over the whole input. \underline{\textit{Temporal validation.}} For each predicted attribute, we take the
peak frame identified in Phase 1, apply a strong blur to it, and check whether the model still
predicts the attribute. We repeat this for randomly selected frames from the same clip. A peak
frame is judged causally important if blurring it changes the prediction more often than
blurring a random frame. We summarise this as a causal score: the ratio of the peak-frame impact
to the average random-frame impact, where a score above $1.0$ means the identified moment
matters more than chance.

\noindent \underline{\textit{Spatial validation.}} We validate the Phase 2 attention regions the
same way. We blur the top $10\%$ highest-attention patches in the peak frame and check whether
the attribute prediction changes, then blur random regions of equal area for comparison. A causal score above $1.0$ indicates the highlighted region is necessary
rather than incidental.

\noindent We use the term \emph{causal} in the sense of a controlled intervention test, blur
the evidence and measure the effect which establishes that the identified evidence is
\emph{needed} for the prediction; it confirms necessity rather than recovering the model's full
internal reasoning.

\begin{figure}[t]
    \centering
    \includegraphics[width=\linewidth]{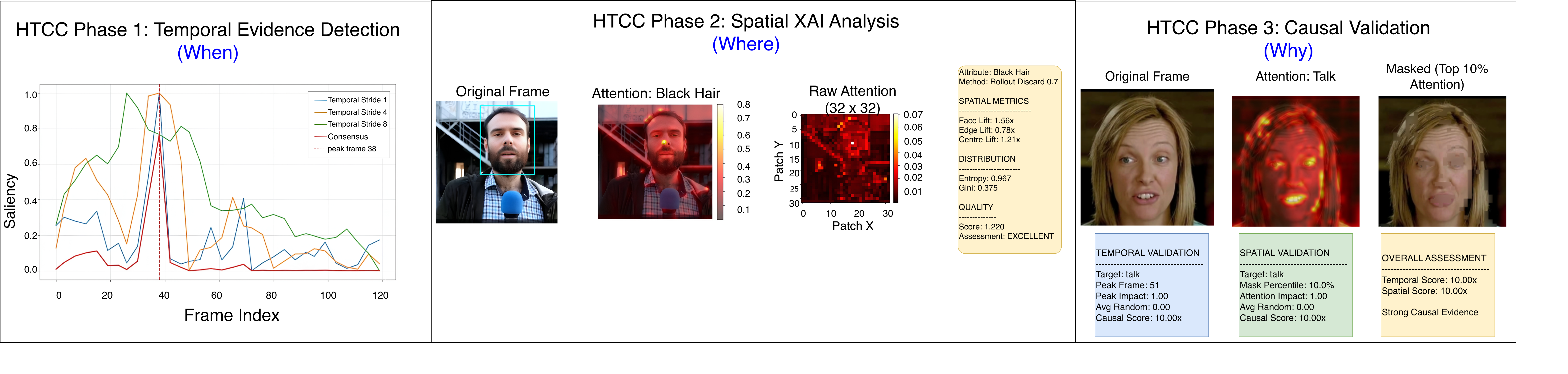}
    \caption{Qualitative H-TCC results. Phase 1 shows the temporal saliency at different time
    windows. Phase 2 shows where the model looks for the attribute \textit{black hair}. Phase 3
    shows the necessity test. }
    \label{fig:qualitative}
    \vspace{-8mm}
\end{figure}

\section{Experiments}
\label{sec:exp}

\noindent \textbf{Tasks, Datasets, and Metrics.} We evaluate VideoFace-X-LLM on six tasks.

\noindent \underline{\textit{Appearance Attributes:}} the model predicts visual attributes such
as facial features, hair, and accessories. This uses CelebV-HQ~\cite{zhu2022celebv} as a
$40$-class multi-label task. We report \textit{Mean Accuracy (mA)} and \textit{Macro F1}. mA
averages per-attribute accuracy; Macro F1 handles the imbalance of rare attributes.

\noindent \underline{\textit{Action Attributes:}} the model predicts facial actions such as
talking, laughing, or singing, on CelebV-HQ~\cite{zhu2022celebv} as a $35$-class multi-label
task. We report mA and Macro F1, as for appearance.

\noindent \underline{\textit{Emotion Recognition:}} we use two settings. At the video level the
model predicts the dominant emotion over the whole clip from eight classes (neutral, happy,
sad, anger, fear, disgust, surprise, contempt). We report accuracy and Macro F1 for both the
video and temporal settings.

\noindent \underline{\textit{Facial Deepfake:}} the model predicts whether a face is real or
manipulated, on FaceForensics++~\cite{rossler2019faceforensics++}. This is binary; we report
accuracy and F1.

\noindent \underline{\textit{Action Units (AUs):}} Action Units are the basic facial-muscle
movements defined by the Facial Action Coding System (FACS)~\cite{ekman1978facial}. We evaluate
on DISFA~\cite{mavadati2013disfa}, which gives frame-level intensity for eight AUs; we treat an
AU as active if intensity $\geq 2$. We report Macro F1 at clip and frame level, plus mA at clip
level.

\noindent \textbf{Note on data.} We fine-tune on CelebV-HQ videos paired with Gemini descriptions. As CelebV-HQ has no official train/test split, we build an identity-independent one by grouping clips by face identity (ArcFace~\cite{deng2019arcface} embeddings) and by source YouTube video and assigning whole groups to one split ($80\%$ train, $20\%$ test), so all CelebV-HQ evaluations are on unseen identities and videos. Deepfake (FaceForensics++) and Action Units (DISFA) are
entirely separate datasets, so they test cross-dataset transfer.

\noindent \textbf{Training Configurations.} We try two settings.
\noindent \underline{\textit{ViT-Only:}} we freeze the language model, and
train only the vision encoder and the connector, so the encoder adapts to faces while the language model stays
fixed.
\noindent \underline{\textit{LLM+ViT:}} we train all parts (encoder, connector, language model)
together.

\noindent \textbf{Implementation Details.} We use the XTuner framework on PyTorch, with $2$
NVIDIA A100 GPUs and FSDP full sharding. Each GPU uses micro-batch size $1$ with gradient
accumulation. We train for $1$ epoch, global batch size $16$, learning rate $4\times10^{-5}$
for all parts, weight decay $0.01$, $3\%$ linear warmup then cosine decay, AdamW with gradient
clipping at norm $1.0$, and $8$ frames per video. For all tasks the model returns structured
JSON, compared against CelebV-HQ ground truth. We compare against off-the-shelf MLLMs
(MiniCPM-V~\cite{yao2024minicpm}, InternVL, Qwen-VL~\cite{yang2025qwen3,bai2025qwen25vl}, LLaVA-NeXT~\cite{zhang2024llava}, Phi-4~\cite{abouelenin2025phi4}) using their public code.

\section{Results}
\label{sec:results}
We report quantitative results, explainability results, and qualitative analysis.

\begin{table*}[t]
\centering
\caption{Results on six tasks. Appearance/Action/Emotion on CelebV-HQ (identity-independent test split, 6,923 clips), Deepfake on FaceForensics++, Action Units on DISFA. Temporal emotion is frame-by-frame at 10\,Hz. mA = Mean Accuracy, F1$_{\text{mac}}$ = Macro F1 (primary metric, robust to imbalance). "Qwen3-VL + EF" = Qwen3-VL fine-tuned on ExplainFace. \textbf{Bold} = best.}
\label{tab:main_results}
\setlength{\tabcolsep}{3pt}
\resizebox{\textwidth}{!}{%
\begin{tabular}{l || cc || cc || cccc || cc || ccc}
\toprule
\multirow{3}{*}{\textbf{Model}} & \multicolumn{2}{c||}{\textbf{Appearance}} & \multicolumn{2}{c||}{\textbf{Action}} & \multicolumn{4}{c||}{\textbf{Emotion}} & \multicolumn{2}{c||}{\textbf{Deepfake}} & \multicolumn{3}{c}{\textbf{Action Units}} \\
& & & & & \multicolumn{2}{c}{\textbf{Video-Level}} & \multicolumn{2}{c||}{\textbf{Temporal}} & & & & \multicolumn{2}{c}{} \\
& mA $\uparrow$ & F1$_{\text{mac}}$ $\uparrow$ & mA $\uparrow$ & F1$_{\text{mac}}$ $\uparrow$ & Acc $\uparrow$ & F1$_{\text{mac}}$ $\uparrow$ & Acc $\uparrow$ & F1$_{\text{mac}}$ $\uparrow$ & Acc $\uparrow$ & F1$_{\text{mac}}$ $\uparrow$ & mA $\uparrow$ & Frame F1 & Clip F1 \\
\hline\hline
InternVideo2.5 (Base)  & 88.10 & 19.48 & 93.69 & 16.44 & 82.06 & 21.96 & 84.04 & 20.04 & 51.01 & 33.96 & 88.3 & 21.1 & 21.6 \\
Ours (ViT-Only)        & 86.16 & 26.29 & 94.50 & 17.88 & 55.61 & 22.00 & 71.47 & 23.83 & 50.00 & 35.23 & 71.0 & 16.7 & 16.8 \\
Ours (ViT+LLM)    & 86.36 & 32.87 & 94.80 & 20.82 & 60.57 & 25.68 & 74.84 & \textbf{26.81} & \textbf{56.32} & 45.80 & 79.2 & 31.6 & 32.2 \\
Qwen3-VL               & \textbf{89.19} & 36.85 & \textbf{95.32} & \textbf{23.22} & 81.09 & 26.11 & 85.88 & 13.06 & 56.07 & \textbf{47.88} & \textbf{88.9} & \textbf{37.7} & \textbf{38.2} \\
Qwen3-VL + EF          & 86.35 & \textbf{39.99} & 94.19 & 23.20 & 78.02 & \textbf{27.47} & 83.47 & 23.28 & 55.71 & 47.63 & 87.9 & 34.3 & 34.7 \\
Qwen2.5-VL             & 87.65 & 28.62 & 95.13 & 18.01 & 81.33 & 24.64 & 85.18 & 21.37 & 50.00 & 33.96 & 86.2 & 23.7 & 24.2 \\
Llava-Next             & 67.91 & 18.45 & 70.22 & 7.88  & \textbf{84.42} & 12.22 & \textbf{86.29} & 9.70 & 50.60 & 33.60 & 11.8 & 18.5 & 18.1 \\
Phi-4                  & 71.99 & 15.60 & 91.68 & 7.53  & 84.41 & 11.70 & 86.18 & 11.66 & 50.00 & 33.33 & 37.0 & 20.4 & 20.0 \\
MiniCPM                & 85.41 & 19.07 & 93.09 & 14.22 & 39.36 & 0.98  & 71.09 & 0.41  & 50.00 & 33.33 & 67.7 & 13.8 & 13.7 \\
\bottomrule
\end{tabular}%
}
\vspace{-8mm}
\end{table*}

\noindent \textbf{Quantitative Results.} \cref{tab:main_results} reports all tasks; every
number below is taken directly from it. \textbf{Appearance.} Fine-tuning on ExplainFace raises Macro F1 from $19.48\%$ to
$32.87\%$ ($+13.39$). The base model's slightly higher mA ($88.10\%$ vs.\ $86.36\%$) comes from
mA being dominated by easy majority-negative attributes; Macro F1, which weights rare
attributes equally, is the reliable measure and improves clearly. \textbf{Action.} Macro F1 rises from $16.44\%$ to $20.82\%$ ($+4.38$), and mA also
improves ($93.69\%\!\rightarrow\!94.80\%$). The ViT-Only setting gains less ($17.88\%$ F1),
which shows language-model adaptation matters for actions. \textbf{Emotion.} Macro F1 rises from $21.96\%$ to $25.68\%$ ($+3.72$) at the video level and from $20.04\%$ to $26.81\%$ ($+6.77$) for temporal emotion, the best temporal F1 of all models. \textbf{Deepfake.} On the held-out FaceForensics++ set we reach $56.32\%$ accuracy, the best of all models, and
$45.80\%$ F1 ($33.96\%\!\rightarrow\!45.80\%$ over the base). \textbf{Action Units.} On held-out DISFA, ViT+LLM raises Macro F1
over base (clip $21.6\%\!\rightarrow\!32.2\%$; frame $21.1\%\!\rightarrow\!31.6\%$) but
lowers mA ($88.3\%\!\rightarrow\!79.2\%$). \textbf{ExplainFace transfers across backbones.} ExplainFace is a transferable
supervision signal, not a single-model trick. Fine-tuning on it raises appearance Macro F1 on
\emph{two} backbones: InternVideo2.5 ($19.48\%\!\rightarrow\!32.87\%$) and Qwen3-VL
($36.85\%\!\rightarrow\!39.99\%$), and it raises Qwen3-VL temporal-emotion Macro F1
($13.06\%\!\rightarrow\!23.28\%$). We do not claim to beat every model on every task: strong
general MLLMs such as Qwen3-VL lead on some tasks. The value of this work is the corpus and the
explainability framework, not a leaderboard win.

\begin{table}[t]
\centering
\caption{Faithfulness of H-TCC explanations. Higher is better for lifts, stability and causal
scores; edge lift should be $<1$; the Adebayo endpoint should fall to about $0$ or below.}
\label{tab:xai_eval}
\scalebox{0.9}{
\begin{tabular}{l c c c}
\toprule
\textbf{Metric} & \textbf{Emotion} & \textbf{Appearance} & \textbf{Action} \\
\hline \hline
Multi-stride stability ($\uparrow$)              & 0.46 & 0.39 & 0.41 \\
Adebayo endpoint ($\rightarrow 0$)               & $-0.12$ & $-0.22$ & $-0.13$ \\
Spatial pointing lift ($\uparrow$, chance $=1$)  & 1.42$\times$ & 1.40$\times$ & 1.40$\times$ \\
Face lift ($\uparrow$, uniform $=1$)             & 1.23$\times$ & 1.23$\times$ & 1.24$\times$ \\
Edge lift ($\downarrow$, want $<1$)              & 0.80$\times$ & 0.80$\times$ & 0.80$\times$ \\
\bottomrule
\end{tabular}}
\vspace{-8mm}
\end{table}

\noindent \textbf{Explainability Results.} \cref{tab:xai_eval} reports the faithfulness of
H-TCC's explanations, and together with the Phase 3 necessity test it answers the central
question: \emph{are the explanations faithful, and is the highlighted evidence actually needed?}
The spatial pointing lift ($1.40$-$1.42\times$) shows the most-attended patch lands on the face
about $1.4\times$ more often than chance, so the model looks at the right place; face lift
($1.23$-$1.24\times$) shows the explanation mass sits inside the face rather than the
background, and edge lift ($0.80<1$) shows it avoids image borders, so it is not
exploiting border artifacts. The Adebayo check drives explanation agreement from
$1.0$ to below $0$ ($-0.12$ to $-0.22$): when the model is randomized the explanations collapse,
so they come from learned knowledge, not a fixed prior. Multi-stride stability ($0.39$-$0.46$)
shows the chosen moments are consistent across time scales. Crucially, the Phase 3 necessity
test then verifies that this evidence \emph{drives} the prediction: blurring the peak frame and
the top-attention region flips the model's decision more often than blurring random frames
or regions, giving temporal ($1.28$-$1.76$) and spatial ($1.44$-$1.93$) causal scores above chance ($1.0$).

\noindent \textbf{Qualitative Analysis.} \cref{fig:qualitative} shows the three
phases on an example. In Phase 1, the saliency is high only in a few short time
windows, not spread over the whole video, so the model picks out the moments that
matter. In Phase 2 (target \textit{black hair}), the model looks at the hair region
and ignores the background. In Phase 3, blurring this evidence drops the prediction; blurring other parts does not.

\section{Conclusion}
We introduce VideoFace-X-LLM, a facial video language model fine-tuned on the weakly
labeled ExplainFace corpus, learning spatio-temporal features that transfer to emotion,
deepfake, Action Unit, and attribute tasks. \textbf{Broader Impact.} It can serve as a base
model for facial video analysis in affective computing, HCI, and behaviour understanding.
\textbf{Limitations.} The model can inherit demographic bias from CelebV-HQ and
preprocessing, and MLLMs can hallucinate; future work will add more diverse data. \textbf{Acknowledgements.} Abhinav Dhall's research was partially supported by the
Australian Research Council under Grant DP260100218.

\bibliographystyle{splncs04}
\bibliography{main}

\end{document}